\documentclass[runningheads]{llncs}

\usepackage[T1]{fontenc}
\usepackage{microtype}

\usepackage{amssymb, amsfonts}
\usepackage{mathtools}
\usepackage{bm}

\usepackage{booktabs}
\usepackage{multirow}
\usepackage{array}

\usepackage{graphicx}
\usepackage{tikz}
\usetikzlibrary{arrows.meta, positioning, patterns, shapes.geometric}
\IfFileExists{animate.sty}{\usepackage{animate}}{}

\usepackage[compatibility=false]{caption}

\definecolor{succblue}{HTML}{2A78D6}
\definecolor{failorange}{HTML}{EB6834}
\definecolor{inkgray}{HTML}{52514E}

\usepackage{xcolor}
\usepackage[colorlinks=true,linkcolor=blue!50!black,
            citecolor=blue!50!black,urlcolor=blue!50!black]{hyperref}
\usepackage{cleveref}
\usepackage{xurl}  
\crefname{section}{Section}{Sections}
\Crefname{section}{Section}{Sections}
\crefname{table}{Table}{Tables}
\crefname{figure}{Figure}{Figures}
\crefname{equation}{Eq.}{Eqs.}

\usepackage{enumitem}
\setlist{itemsep=2pt,topsep=2pt,parsep=0pt,partopsep=0pt}

\newcommand{\E}{\mathbb{E}}
\newcommand{\task}[1]{\textit{#1}}
\newcommand{\figfile}[1]{%
  \IfFileExists{#1.pdf}{\includegraphics[width=\textwidth]{#1.pdf}}{%
  \IfFileExists{#1.jpg}{\includegraphics[width=\textwidth]{#1.jpg}}{%
  \fbox{\parbox{0.92\textwidth}{\centering\vspace{1.2cm}\texttt{\detokenize{#1}}.pdf / .jpg not found\vspace{1.2cm}}}}}}

\title{Practice Is Not Free: Measuring the Embodied Cost of Reset-Free Robot Practice}
\titlerunning{Practice Is Not Free}
\author{Vedant Telrandhe\inst{1} \and Sushil Sharma\inst{2}}
\authorrunning{Telrandhe and Sharma}
\institute{Independent Researcher\\
  \email{vedantm264@gmail.com}
  \and
  \text{University of Limerick}\\
  \email{sushil.sharma@ul.ie}}

\begin{document}
\maketitle

\begin{center}
\ifdefined\animategraphics
  \IfFileExists{fig1_run_anim.pdf}%
    {\animategraphics[width=\textwidth,autoplay,loop,poster=last]{4}{fig1_run_anim}{}{}}%
    {\figfile{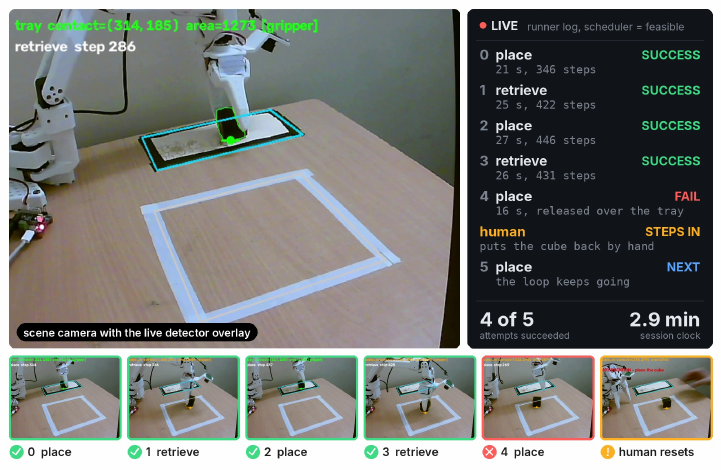}}%
\else
  \figfile{fig1_run}%
\fi
\captionof{figure}{\textbf{One recorded run with nobody resetting the scene: four successes, one failure, and one call for help.}
\emph{Left:} the scene camera with the detector's live overlay; the cyan box is the tray, the outline and dot are the cube and its contact point, and the label turns green when the cube is in the tray. \emph{Right:} a summary of the runner's log for the same run. \emph{Bottom:} one frame per attempt. \task{Place} and \task{retrieve} undo each other until attempt~4 releases over the tray without the cube and a person resets it. This run was recorded separately and is not one of the five sessions analyzed in \cref{sec:results}; videos of the loop are in the repository.\label{fig:teaser}}
\end{center}
\vspace{-0.6\baselineskip}
\medskip\noindent\textbf{Code, data, logs and videos:} \url{https://github.com/Vedant2604/so101-cost-aware-scheduler}

\begin{abstract}
When a robot practices without human resets, every attempt still costs wall-clock time, actuator wear, and sometimes a person's attention, yet the schedulers that decide what to practice next treat attempts as free. We present an open, low-cost testbed for measuring these costs: a self-built SO-101 arm runs two demonstration-trained ACT policies that reset each other, with a vision-based success detector and a ledger that records the time, joint travel, and human interventions of every attempt. We formulate cost-aware practice scheduling, in which task selection is rewarded by measured competence gain net of embodied cost, and specify in advance the comparison that will test it. Its first two conditions, run on hardware for 51 minutes of robot time, show that a fixed retry limit did not reduce the share of execution time lost to failures (36\%, against 27\% for a feasibility rule; not significant) and saved no robot time. The logs suggest why: failures showed no sign of clustering, so a retry right after a failure was as likely to succeed as any other attempt. The ledger also shows that most of the retry rule's extra human time came from cubes knocked out of reach rather than from the rule itself, and a cost-aware rule replayed on the logs recovers the sensible choice without a hand-set limit. We also report lessons from building the system, including a coverage effect in which concentrating demonstrations in a smaller workspace raised success there from 8/20 to 10/10.

\keywords{Reset-free learning \and Autonomous practice \and Embodied cost \and Task scheduling \and Low-cost robotics.}

\end{abstract}

\section{Introduction}
\label{sec:intro}

A robot that keeps improving after deployment must decide, again and again, what to practice next. On physical hardware that decision is never free. Each attempt costs wall-clock time and actuator wear, and an attempt that leaves the scene in a state the robot cannot recover from costs a person's attention. This paper argues that these costs should enter the practice decision directly, and presents a low-cost testbed built to measure them (\cref{fig:teaser}).

Reset-free reinforcement learning was motivated by exactly this concern. Real-world learning stalls when a human must reset the scene after every attempt~\cite{eysenbach2018,zhu2020}. The field has largely addressed it by making resets automatic: learning a reset policy alongside the forward policy~\cite{eysenbach2018}, or choosing tasks so that each one resets another~\cite{gupta2021}. These methods reduce the number of human resets. They do not, however, price the attempts the robot makes on its own. Eysenbach et al.~\cite{eysenbach2018} note that the abort threshold should depend on the cost of a hard reset.

A second line of work asks which skill a robot should practice. Estimate, Extrapolate, Situate (EES) selects the skill whose predicted competence gain most improves task success~\cite{kumar2024}. Deliberate Practice computes a budget-optimal allocation of practice across skills~\cite{vats2026}. Both are real advances, and both count the practice budget in attempts. An attempt that takes 18 seconds and one that takes 48 seconds, or one that ends in a human intervention and one that does not, count the same.

Our own system made this concrete. In its first unattended session, a hand-designed scheduler alternated two policies that reset each other. Late in the session, one policy began to fail from the same start state. The scheduler retried it at the full episode cost of 48 seconds and a complete arm excursion, and would have continued indefinitely, because nothing in its decision rule registers that practice is expensive.

This paper makes four contributions:
\begin{enumerate}
\item \textbf{A formulation of cost-aware practice scheduling.} The scheduler is rewarded by the measured competence gain of the low-level policies, net of embodied cost: wall-clock time, joint travel as a proxy for actuator wear, and human interventions (\cref{sec:problem}).
\item \textbf{An open, low-cost testbed.} A self-built SO-101 arm runs two demonstration-trained ACT policies~\cite{zhao2023} that reset each other, with a vision-based success detector and a per-episode cost ledger. The code, datasets, and trained policies are public (\cref{sec:system}).
\item \textbf{Lessons from building it.} Validation loss that rose while real-world success improved, a coverage effect in demonstration placement, a reset-to-home requirement absent from the demonstrations, and a data-encoding fix that cut training from 50 hours to 12 (\cref{sec:pilot}).
\item \textbf{An evaluation protocol and its first results.} We specify a comparison of cost-blind and cost-aware schedulers on identical hardware before running it, and report its first two conditions, a feasibility rule and a fixed retry limit; the cost-aware conditions are the next experiment (\cref{sec:eval}).
\end{enumerate}

\section{Related Work}
\label{sec:related}

Prior work has made practice autonomous and has learned which skill to practice. Other work budgets human interventions. To our knowledge, none weighs the full physical cost of each practice attempt, in robot time, wear and human attention, when choosing what to practice next on hardware.

\paragraph{Reset-free and autonomous reinforcement learning.}
Leave No Trace learns a reset policy jointly with the forward policy and aborts episodes predicted to become irreversible~\cite{eysenbach2018}. Zhu et al.~\cite{zhu2020} identify the absence of resets as one of three requirements for real-world robotic RL. Gupta et al.~\cite{gupta2021} observe that a well-chosen set of tasks can reset one another, and train a dexterous hand for over 60 hours without human resets. VaPRL generates a curriculum of start states that reduces the number of human interventions relative to episodic RL in simulation~\cite{sharma2021}. EARL formalizes the setting as a benchmark~\cite{sharma2022}. DBAP uses demonstrations to bootstrap both the low-level policies and the sequencing of tasks for a real kitchen environment~\cite{gupta2023}. PAINT learns to recognize irreversible states and to request help only when needed~\cite{xie2022}. In interactive imitation learning, ThriftyDAgger requests human help within a budget on interventions~\cite{hoque2021}, and Fleet-DAgger allocates one person's attention across a fleet of robots~\cite{hoque2022}. Across this line of work, human interventions are treated as an outcome to be minimized or budgeted. The time and wear consumed by each autonomous attempt do not enter the decision of what to attempt next.

\paragraph{Choosing what to practice.}
Learning progress has long been used to select practice. Stout and Barto~\cite{stout2010} and Baranes and Oudeyer~\cite{baranes2013} direct exploration toward goals where competence is improving. Graves et al.~\cite{graves2017} cast task selection as a nonstationary multi-armed bandit rewarded by learning progress. EES brings this to real robots, estimating each skill's competence and extrapolating how practice would change it~\cite{kumar2024}. Deliberate Practice shows that such greedy selection can be myopic, and computes a budget-optimal allocation with a bilinear program~\cite{vats2026}. FIND uses a vision-language agent to pick practice tasks for a frozen VLA, favoring those with low recent success~\cite{fang2026}. All three measure practice in attempts: EES prices a skill at $-\log(\text{competence})$, Deliberate Practice caps the number of trials, and FIND ranks by recent success alone. Our formulation keeps the competence-driven objective and replaces the unit-cost attempt with its measured embodied cost.

\paragraph{Resets in simulation.}
OmniReset shows that when resets are free, as in a simulator, the right strategy is to generate them as diversely as possible and let compute buy behavioral coverage; the resulting policies transfer zero-shot and retry robustly on real hardware~\cite{yin2026}. Our setting is the complement. Once a policy is deployed and must keep improving in place, every reset is physical, and the question becomes which attempts are worth their cost.

\section{Problem Formulation}
\label{sec:problem}

We treat practice scheduling as a sequential decision made once per episode, in which every option has a measured physical price.

\paragraph{Setting.}
The robot holds a set of $K$ tasks, each with a low-level policy that may or may not be improving. Before episode $n$, the scheduler observes a coarse scene state $s_n$ (in our system, which region the object occupies) and selects an action $a_n$. Each episode returns a binary success label $y_n$ and a cost vector
\begin{equation}
  c_n = \bigl( t_n,\; d_n,\; h_n \bigr), \qquad
  d_n = \sum_{\tau} \sum_{j} \bigl| q_{\tau, j} - q_{\tau-1, j} \bigr|,
  \label{eq:cost}
\end{equation}
where $t_n$ is wall-clock time, $d_n$ is summed absolute joint displacement over the episode (a proxy for actuator wear), and $h_n$ is the number of human interventions.

\paragraph{Actions.}
The action set is what makes scheduling a decision rather than a lookup. With only two mutually resetting tasks and a feasibility filter, each scene state admits exactly one legal task; this is the regime of our pilot session, and in it the scheduler makes no real choice. We therefore define three kinds of action: \emph{attempt} a task whose start condition holds; \emph{reposition} the object within its region without completing any task, which changes the start state at a cost; and \emph{request help}, which restores a known-good state at a cost of one intervention. Requesting help as a priced action generalizes the fixed rule in Leave No Trace, which gives the reset policy $N$ attempts before a manual reset is needed, and notes that the choice of $N$ should depend on the cost of a hard reset~\cite{eysenbach2018}.

\paragraph{Objective.}
Let $G_n$ be the competence gain attributable to episode $n$, estimated online from success labels in the style of learning-progress curricula~\cite{graves2017,kumar2024}. The cost-aware scheduler $\sigma$ maximizes gain net of a weighted cost:
\begin{equation}
  J(\sigma) = \E_{\sigma}\Bigl[\, \sum_{n=1}^{N} \bigl( G_n - \lambda^{\top} c_n \bigr) \Bigr],
  \qquad \lambda = \bigl( \lambda_t,\; \lambda_d,\; \lambda_h \bigr) \geq 0.
  \label{eq:objective}
\end{equation}
The weights $\lambda$ express how much an operator values a second of robot time, a unit of joint travel, and each interruption of their own attention. Setting $\lambda = 0$ recovers a cost-blind learning-progress scheduler. Setting $G_n = 0$ recovers a scheduler that only minimizes cost.

\paragraph{Two regimes.}
When the low-level policies are frozen, as in our current system, competence does not change with practice. $G_n$ is then replaced by the success label $y_n$, and the objective trades successes against their embodied cost: the scheduler should stop paying for attempts that are very unlikely to succeed. When the policies are updated from practice data, $G_n$ measures genuine improvement, and the scheduler must trade the expected learning value of an attempt against its price. The first regime is testable on the current platform. The second requires closing the loop from practice data back to the policies, which we list as future work in \cref{sec:future}.

\section{System}
\label{sec:system}

\begin{figure}[t]
\centering
\figfile{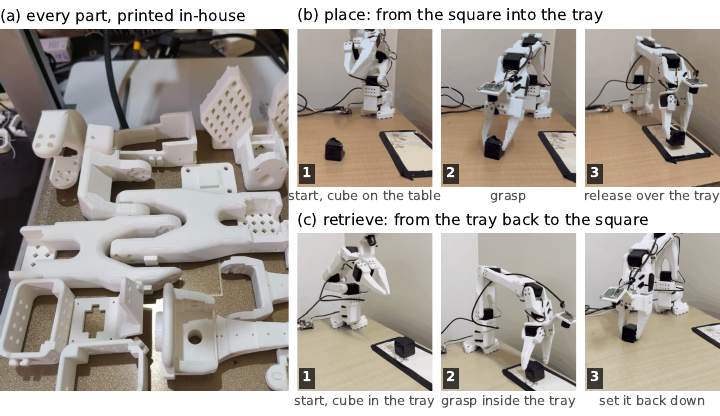}
\caption{\textbf{A robot we printed and assembled ourselves, doing both tasks.} (a) Every structural part of the SO-101, printed in-house before assembly. (b) \task{Place}: the arm starts at its home pose, grasps the cube in the taped square, and releases it over the tray. (c) \task{Retrieve}: the same arm reaches into the tray, grasps the cube, and sets it back in the square, which is exactly where \task{place} begins.}
\label{fig:testbed}
\end{figure}

\begin{figure}[t]
\centering
\figfile{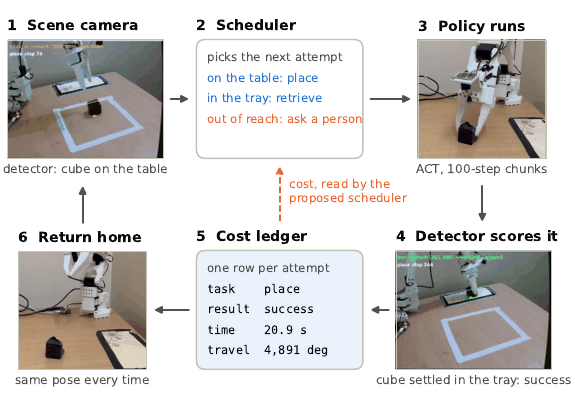}
\caption{\textbf{The practice loop, drawn with real frames from one unattended cycle.} (1) The scene camera sees the cube, and the detector marks it as on the table. (2) The scheduler turns that into the next attempt: \task{place} from the table, \task{retrieve} from the tray, a call for help if the cube is out of reach. (3) The chosen ACT policy drives the arm. (4) The detector scores the result; here the cube has settled in the tray. (5) The ledger writes one row for the attempt; the values shown are a real row from our logs. (6) The arm returns to the pose every demonstration started from, and the loop repeats. The dashed arrow is what this paper adds: today's scheduler records the cost but never reads it, and the proposed scheduler decides with it.}
\label{fig:progress}
\end{figure}

The testbed is a single low-cost arm that practices two tasks unattended and records what every attempt costs (\cref{fig:testbed,fig:progress}).

\paragraph{Hardware.}
The arm is an SO-101, a six-servo open-source manipulator supported by the LeRobot library~\cite{cadene2026}, which we assembled from 3D-printed parts and Feetech serial-bus servos. Two 720p webcams, one at the wrist and one at the side, provide observations, and demonstrations were collected with a PS5 gamepad. Inference runs on a 6\,GB RTX 2060; training used a rented RTX 5090.

\paragraph{Tasks and policies.}
The workspace has a taped region on the table and a tray. Two tasks reset each other, in the pattern of Gupta et al.~\cite{gupta2021}: \task{place} moves the cube from the table region to the tray, and \task{retrieve} returns it. Each task has its own action-chunking transformer (ACT)~\cite{zhao2023}. Ours is a 52M-parameter instance with a ResNet-18 backbone, predicting chunks of 100 future actions; the original is larger, at about 80M parameters. \task{Place} was trained on 43 teleoperated demonstrations inside the taped region. \task{Retrieve} was first trained on 51 demonstrations (the pilot session, \cref{sec:pilot-session}) and later retrained with 25 additional demonstrations that start from failure states (76 in total), which is the version used in \cref{sec:results}. We use one policy per task: \task{place} and \task{retrieve} look alike mid-carry but need opposite motions, and a single multi-task policy trained on both suffered from that interference.

\paragraph{Success detection.}
A blob detector on the scene camera classifies the cube as on the table, in the tray, or outside both. It took three corrections to make it reliable. Under perspective, a cube resting in the tray has its centroid outside the tray's outline, so the region is assigned from the blob's contact point instead. A cube carried through the air can also look as if it is in the tray, so success requires it to stay within 6\,px for 15 frames. And because the arm hides the tray at the moment of release, a \task{place} also counts when the gripper opens inside the base-angle window learned from the demonstrations and the cube has left the table.

\paragraph{Scheduler.}
The baseline is a hand-designed feasibility scheduler. If the cube is in the table region it selects \task{place}; if it is in the tray it selects \task{retrieve}; if it is outside both, it requests human help and logs an intervention.

\paragraph{Homing.}
Between episodes the arm returns to a fixed home pose over about 5\,s. This step was not in the original design; \cref{sec:observations} describes why it became necessary.

\paragraph{Cost ledger.}
Each episode is capped at 800 control steps, about 48\,s. For every episode the runner writes one row: task, start region, success, steps, seconds, summed joint travel, and cumulative interventions. The ledger is the measurement instrument for everything in \cref{sec:problem,sec:eval}.

\paragraph{Availability.}
The code, session logs, analysis scripts, demonstration datasets, and trained policies are public at \url{https://github.com/Vedant2604/so101-cost-aware-scheduler}, together with videos of the autonomous loop. Every photographic and chart figure is regenerated from those logs and videos by \texttt{paper/make\_figures.py} and \texttt{paper/make\_teaser.py}, and the offline analyses of \cref{sec:results} by \texttt{analyze\_offline.py}.

\section{Pilot Session and Lessons from Building}
\label{sec:pilot}

\subsection{First logged session}
\label{sec:pilot-session}

The first unattended session used the feasibility scheduler, ran for ten minutes, and completed 11 episodes. Six succeeded (\task{place} 3 of 5, \task{retrieve} 3 of 6) and a person was needed once. Episodes 5 to 9 ran two complete place-and-retrieve cycles with nobody watching. The price was visible at once: a successful episode took 18 to 40\,s, while the failed attempts consumed 242 of the 434\,s of policy execution, or 56\%. The full ledger is \texttt{runs/pilot.csv} in the repository.

The end of the session is what motivates this paper. \task{Retrieve} failed twice in succession from the same state, each time hovering 1 to 2\,cm above the cube without descending far enough to grasp it. The feasibility scheduler, seeing the cube in the tray, selected \task{retrieve} again, and would have kept doing so at 48\,s and a full arm excursion per attempt, because it had no other action to take.

\subsection{Lessons from building the system}
\label{sec:observations}

\paragraph{For a narrow task, the robot should pick the checkpoint, not the loss curve.}
For the initial pick-and-place policy, held-out validation loss rose from 0.26 to 0.30 between 20k and 70k training steps, the usual sign of overfitting. Over the same interval the policy went from approaching without grasping (20k) to grasping unreliably (40k) to completing the task unaided (60k). With the same cube, workspace and tray every time, fitting this setup's trajectories more closely raises the loss on unseen demonstrations while improving behavior in the only setup the policy will run in.

\paragraph{Concentrating demonstrations beat adding them.}
A \task{place} policy trained on 43 demonstrations recorded inside a small taped region succeeded in 10 of 10 trials there, against 8 of 20 for an earlier policy trained on 51 demonstrations spread over a wider area. This suggests that density of coverage, more than the number of demonstrations, was the limit; these were standalone trials from hand-placed starts, and in the closed loop of \cref{sec:results} the same policy succeeded on 28 of 47 attempts.

\paragraph{Autonomous practice needed a reset the demonstrations never showed.}
Run on its own, the \task{place} policy succeeded first time; inside the loop it failed 5 of 5 times, because each episode began wherever the last one ended while every demonstration began from the home pose. Returning to that pose between episodes removed this failure mode.

\paragraph{Data encoding, not compute, set the training budget.}
Stored as individual images, the camera frames left the GPU idle at 4 samples per second. Re-encoding the same data as video cut the dataset from 4.1\,GB to 212\,MB, raised throughput to 44 samples per second, and turned a 50-hour training run into a 12-hour one.

\section{Proposed Scheduler and Evaluation Protocol}
\label{sec:eval}

The test of this paper's claim is a controlled comparison of four schedulers on the same arm, the same policies, and the same session length. We specified it before running it; \cref{sec:results} reports the first two conditions.

\subsection{A cost-aware scheduler}
\label{sec:scheduler}

The scheduler needs a real choice, so we first extend the action set. A third policy, \task{reposition}, will be trained from demonstrations to move the cube within its current region without completing either task. It gives the scheduler a way to spend an episode changing the start state rather than retrying from a bad one. \task{Request help} becomes an explicit action whose cost is one intervention.

The scheduler then treats action selection as a nonstationary multi-armed bandit, following Graves et al.~\cite{graves2017}. For each action and coarse start region it maintains a sliding-window estimate of success probability, in the spirit of the competence estimates of EES~\cite{kumar2024}, and a running mean of each component of the ledger cost. Each episode is rewarded by the success it produced, or later by competence gain, minus the $\lambda$-weighted cost of \cref{eq:objective}. The sliding window lets the estimates follow a policy whose performance drifts, as \task{retrieve}'s did at the end of the pilot session.

The intended behavior in the pilot's final state is concrete. After two failed retrieves from the same state, the estimated success of a third attempt falls while its cost stays at 48\,s and a full excursion. A reposition, or for a high enough $\lambda_h$ a request for help, becomes the better purchase.

\subsection{Conditions}
\label{sec:conditions}

\begin{table}[t]
\centering
\small
\setlength{\tabcolsep}{4pt}
\caption{\textbf{Four schedulers on the same arm, from no choice to a priced choice.} A and B are the rule-based baselines reported in \cref{sec:results}; C adds a learned schedule without prices; D is the proposed method.}
\label{tab:conditions}
\begin{tabular}{l>{\raggedright}p{2.6cm}l>{\raggedright\arraybackslash}p{3.8cm}}
\toprule
\textbf{Condition} & \textbf{Action set} & \textbf{Reward} & \textbf{Role} \\
\midrule
A. Feasibility        & attempt only                      & none                     & the pilot baseline \\
B. Fixed retry limit  & attempt, request help             & none                     & requests help after $N$ consecutive failures, a fixed reset budget in the spirit of Leave No Trace \\
C. Cost-blind bandit  & attempt, reposition, request help & success, $\lambda = 0$   & isolates the effect of a learned schedule \\
D. Cost-aware bandit  & attempt, reposition, request help & success $-\lambda^{\top}c$ & the proposed method \\
\bottomrule
\end{tabular}
\end{table}

\Cref{tab:conditions} lists the conditions. Condition B matters most. If a hand-set retry limit performs as well as the cost-aware bandit, then pricing practice adds nothing over a sensible rule, and we will report that result as found.

\subsection{Protocol and metrics}
\label{sec:protocol}

Each session lasts 10 minutes, with 20 sessions per condition in randomized blocks to guard against drift from servo heating, lighting and wear. Policies are frozen, placing the experiment in the first regime of \cref{sec:problem}. $\lambda$ is fixed before data collection, and a second sweep varies $\lambda_h$ alone to test whether behavior shifts as human attention becomes more expensive.

The primary metrics are successes per minute, successes per unit of joint travel, and interventions per success; the share of execution time lost to failures is secondary. All are reported with 95\% intervals over sessions. The first results in \cref{sec:results} use Wilson intervals over attempts, because five sessions are too few for session-level intervals.

We state three hypotheses in advance. \textbf{H1:} condition D spends a smaller fraction of execution time on failed attempts than condition A. \textbf{H2:} condition D achieves more successes per unit embodied cost than condition C. \textbf{H3:} condition D matches or improves on condition B in interventions per success, without a hand-tuned retry limit.

\subsection{First results: conditions A and B}
\label{sec:results}

Over 51 minutes of robot time, the feasibility rule (A) succeeded on 79\% of attempts and spent 27\% of execution time on failures. The retry limit with $N = 2$ (B) succeeded on 66\%, spent 36\%, and asked for help twice as often per success.

\paragraph{Setup.}
We ran 5 sessions of about 10 minutes of robot time each, 3 under A and 2 under B, in the order A, B, B, A, A on 12 and 13 October 2026. Both conditions used the frozen \task{place} and \task{retrieve} policies of \cref{sec:system}. Two changes to the protocol of \cref{sec:protocol} apply. First, the session clock pauses while a person is helping, so both conditions receive the same robot time and human time is logged as a separate cost. Three earlier sessions run without this pause are excluded. Second, attempts shorter than 2\,s are invalid, because no policy completes a task that fast. Both such attempts occurred in one B session, where the cube was still being moved by hand as the episode began. The 2\,s threshold was set after these were observed; the effect of including them is reported under Statistical scope below.

\begin{table}[t]
\centering
\small
\setlength{\tabcolsep}{5pt}
\caption{\textbf{Conditions A and B, pooled over sessions.} Confidence intervals are Wilson score intervals over attempts; they treat attempts as independent and are therefore optimistic. $^{*}$Execution time excludes homing and detection, which \cref{fig:ledger} counts separately.}
\label{tab:ab}
\begin{tabular}{lcc}
\toprule
\textbf{Metric} & \textbf{A. Feasibility} & \textbf{B. Retry limit ($N=2$)} \\
\midrule
Sessions, robot minutes                    & 3, 30.9        & 2, 20.1 \\
Valid attempts                             & 47             & 32 \\
Success rate (95\% CI)                     & 78.7\% (65.1--88.0) & 65.6\% (48.3--79.6) \\
Place, retrieve successes                  & 17/26, 20/21   & 11/21, 10/11 \\
Execution time on failed attempts$^{*}$    & 26.6\%         & 35.7\% \\
Successes per robot minute                 & 1.20           & 1.05 \\
Execution time per success (s)             & 36.9           & 41.8 \\
Joint travel per success (deg, all joints) & 7{,}690        & 8{,}326 \\
Interventions per success                  & 0.14 (5 total) & 0.29 (6; 3 retry-limit stops) \\
Human time per success (s)                 & 2.3            & 10.4 \\
Longest run of consecutive failures        & 2              & 3 \\
\bottomrule
\end{tabular}
\end{table}

\begin{figure}[t]
\centering
\figfile{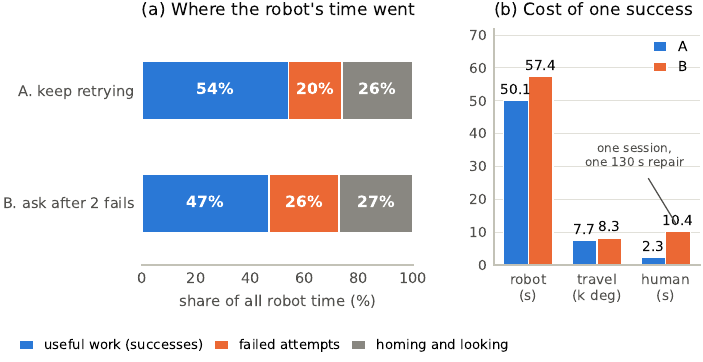}
\caption{\textbf{A quarter of the robot's time goes to neither success nor failure.} (a) How every minute of robot time was spent under each rule. Blue is useful work, orange is failed attempts, and gray is homing and waiting for the detector, an overhead paid on every attempt whichever rule chooses it. (b) What one success cost in each of the ledger's three currencies. Robot time here includes homing, so it is higher than the execution time in \cref{tab:ab}. Most of rule B's human time came from cubes knocked out of reach, including a single 130\,s repair, not from the retry rule itself (\cref{sec:results}).}
\label{fig:ledger}
\end{figure}

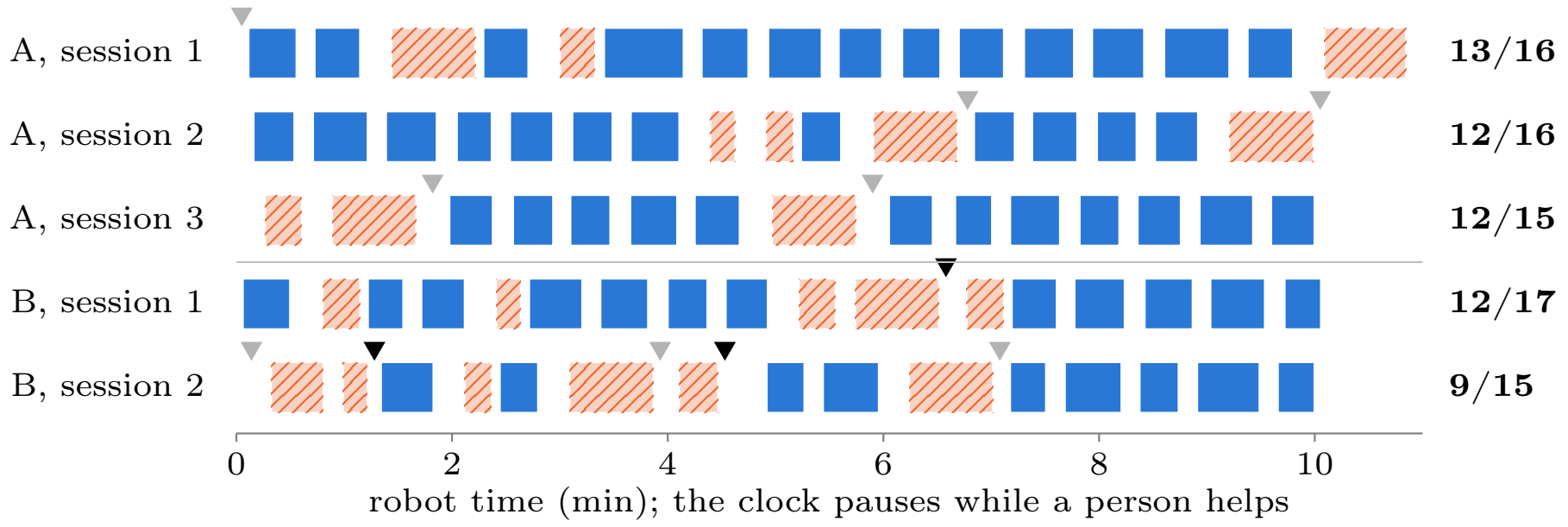
\begin{figure}[t]
\centering
\resizebox{\textwidth}{!}{%
\begin{tikzpicture}[x=0.80cm,y=0.62cm,font=\scriptsize,
    succ/.style={fill=succblue, draw=white, line width=0.3pt},
    fail/.style={fill=failorange!30, draw=white, line width=0.3pt,
                 postaction={pattern=north east lines, pattern color=failorange}},
    stop/.style={isosceles triangle, isosceles triangle apex angle=60, shape border rotate=-90,
                 fill=black, inner sep=0pt, minimum size=4pt},
    help/.style={stop, fill=gray!60}]
\node[help] at (0.050,4.47) {};
\draw[succ] (0.115,3.70) rectangle (0.555,4.30);
\draw[succ] (0.728,3.70) rectangle (1.145,4.30);
\draw[fail] (1.440,3.70) rectangle (2.215,4.30);
\draw[succ] (2.293,3.70) rectangle (2.705,4.30);
\draw[fail] (3.002,3.70) rectangle (3.325,4.30);
\draw[succ] (3.412,3.70) rectangle (4.145,4.30);
\draw[succ] (4.317,3.70) rectangle (4.745,4.30);
\draw[succ] (4.935,3.70) rectangle (5.425,4.30);
\draw[succ] (5.588,3.70) rectangle (5.985,4.30);
\draw[succ] (6.177,3.70) rectangle (6.525,4.30);
\draw[succ] (6.702,3.70) rectangle (7.115,4.30);
\draw[succ] (7.307,3.70) rectangle (7.765,4.30);
\draw[succ] (7.938,3.70) rectangle (8.415,4.30);
\draw[succ] (8.607,3.70) rectangle (9.205,4.30);
\draw[succ] (9.380,3.70) rectangle (9.795,4.30);
\draw[fail] (10.092,3.70) rectangle (10.845,4.30);
\node[anchor=east] at (-0.15,4) {A, session 1};
\node[anchor=west] at (11.1,4) {\textbf{13/16}};
\draw[succ] (0.162,2.70) rectangle (0.535,3.30);
\draw[succ] (0.713,2.70) rectangle (1.215,3.30);
\draw[succ] (1.390,2.70) rectangle (1.855,3.30);
\draw[succ] (2.047,2.70) rectangle (2.365,3.30);
\draw[succ] (2.540,2.70) rectangle (2.935,3.30);
\draw[succ] (3.118,2.70) rectangle (3.485,3.30);
\draw[succ] (3.660,2.70) rectangle (4.105,3.30);
\draw[fail] (4.397,2.70) rectangle (4.625,3.30);
\draw[fail] (4.915,2.70) rectangle (5.165,3.30);
\draw[succ] (5.238,2.70) rectangle (5.605,3.30);
\draw[fail] (5.910,2.70) rectangle (6.685,3.30);
\node[help] at (6.780,3.47) {};
\draw[succ] (6.843,2.70) rectangle (7.215,3.30);
\draw[succ] (7.382,2.70) rectangle (7.795,3.30);
\draw[succ] (7.983,2.70) rectangle (8.345,3.30);
\draw[succ] (8.522,2.70) rectangle (8.915,3.30);
\draw[fail] (9.210,2.70) rectangle (9.985,3.30);
\node[help] at (10.050,3.47) {};
\node[anchor=east] at (-0.15,3) {A, session 2};
\node[anchor=west] at (11.1,3) {\textbf{12/16}};
\draw[fail] (0.267,1.70) rectangle (0.605,2.30);
\draw[fail] (0.890,1.70) rectangle (1.665,2.30);
\node[help] at (1.820,2.47) {};
\draw[succ] (1.978,1.70) rectangle (2.375,2.30);
\draw[succ] (2.568,1.70) rectangle (2.935,2.30);
\draw[succ] (3.100,1.70) rectangle (3.465,2.30);
\draw[succ] (3.655,1.70) rectangle (4.085,2.30);
\draw[succ] (4.253,1.70) rectangle (4.665,2.30);
\draw[fail] (4.970,1.70) rectangle (5.745,2.30);
\node[help] at (5.900,2.47) {};
\draw[succ] (6.055,1.70) rectangle (6.455,2.30);
\draw[succ] (6.668,1.70) rectangle (7.005,2.30);
\draw[succ] (7.178,1.70) rectangle (7.635,2.30);
\draw[succ] (7.822,1.70) rectangle (8.185,2.30);
\draw[succ] (8.360,1.70) rectangle (8.755,2.30);
\draw[succ] (8.935,1.70) rectangle (9.425,2.30);
\draw[succ] (9.598,1.70) rectangle (9.995,2.30);
\node[anchor=east] at (-0.15,2) {A, session 3};
\node[anchor=west] at (11.1,2) {\textbf{12/15}};
\draw[succ] (0.062,0.70) rectangle (0.495,1.30);
\draw[fail] (0.803,0.70) rectangle (1.145,1.30);
\draw[succ] (1.222,0.70) rectangle (1.545,1.30);
\draw[succ] (1.718,0.70) rectangle (2.115,1.30);
\draw[fail] (2.410,0.70) rectangle (2.635,1.30);
\draw[succ] (2.717,0.70) rectangle (3.205,1.30);
\draw[succ] (3.378,0.70) rectangle (3.815,1.30);
\draw[succ] (4.003,0.70) rectangle (4.365,1.30);
\draw[succ] (4.540,0.70) rectangle (4.925,1.30);
\draw[fail] (5.217,0.70) rectangle (5.555,1.30);
\draw[fail] (5.740,0.70) rectangle (6.515,1.30);
\node[stop] at (6.580,1.47) {};
\draw[fail] (6.773,0.70) rectangle (7.115,1.30);
\draw[succ] (7.193,0.70) rectangle (7.605,1.30);
\draw[succ] (7.775,0.70) rectangle (8.235,1.30);
\draw[succ] (8.425,0.70) rectangle (8.865,1.30);
\draw[succ] (9.035,0.70) rectangle (9.535,1.30);
\draw[succ] (9.723,0.70) rectangle (10.055,1.30);
\node[anchor=east] at (-0.15,1) {B, session 1};
\node[anchor=west] at (11.1,1) {\textbf{12/17}};
\node[help] at (0.140,0.47) {};
\draw[fail] (0.325,-0.30) rectangle (0.805,0.30);
\draw[fail] (0.990,-0.30) rectangle (1.215,0.30);
\node[stop] at (1.280,0.47) {};
\draw[succ] (1.343,-0.30) rectangle (1.825,0.30);
\draw[fail] (2.117,-0.30) rectangle (2.365,0.30);
\draw[succ] (2.445,-0.30) rectangle (2.795,0.30);
\draw[fail] (3.090,-0.30) rectangle (3.865,0.30);
\node[help] at (3.930,0.47) {};
\draw[fail] (4.108,-0.30) rectangle (4.465,0.30);
\node[stop] at (4.530,0.47) {};
\draw[succ] (4.920,-0.30) rectangle (5.265,0.30);
\draw[succ] (5.442,-0.30) rectangle (5.955,0.30);
\draw[fail] (6.240,-0.30) rectangle (7.015,0.30);
\node[help] at (7.080,0.47) {};
\draw[succ] (7.177,-0.30) rectangle (7.505,0.30);
\draw[succ] (7.685,-0.30) rectangle (8.205,0.30);
\draw[succ] (8.378,-0.30) rectangle (8.735,0.30);
\draw[succ] (8.912,-0.30) rectangle (9.485,0.30);
\draw[succ] (9.660,-0.30) rectangle (9.995,0.30);
\node[anchor=east] at (-0.15,0) {B, session 2};
\node[anchor=west] at (11.1,0) {\textbf{9/15}};
\draw[gray!50] (0,1.5) -- (11,1.5);
\draw[gray] (0,-0.55) -- (11,-0.55);
\foreach \x in {0,2,4,6,8,10} {\draw[gray] (\x,-0.55) -- (\x,-0.65); \node[below] at (\x,-0.6) {\x};}
\node[below] at (5.5,-1.0) {robot time (min); the clock pauses while a person helps};
\end{tikzpicture}
}
\caption{\textbf{Every attempt in all five sessions, in the order it happened.} Each block is one attempt and its width is how long it took; blue succeeded and hatched orange failed. Black triangles mark the retry limit stopping to ask for help; gray triangles mark a person stepping in, either to set up the cube at the start of a session or to move one that was out of reach. The number at the right of each row is that session's successes. Long runs of orange are rare, which is why a rule that counts failures had little to act on.}
\label{fig:timeline}
\end{figure}

\begin{figure}[t]
\centering
\figfile{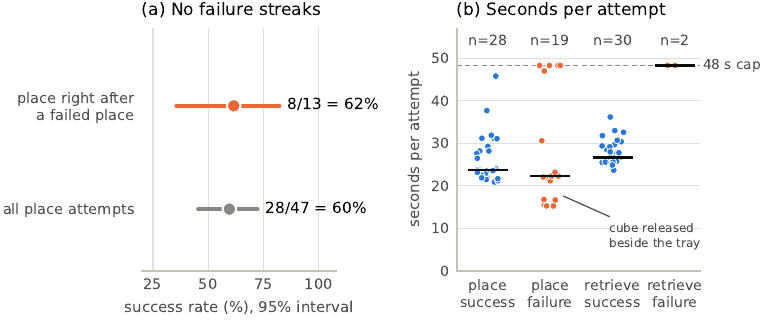}
\caption{\textbf{A failure does not predict another, and most failures are cheap.} (a) A \task{place} attempt straight after a failed \task{place} succeeded 62\% of the time, the same as \task{place} attempts overall; bars are 95\% Wilson intervals. (b) The duration of every valid attempt, with black lines at the medians and the 800-step cap dashed. Most failed \task{place} attempts end early, because the gripper releases over the tray and the cube lands beside it; the expensive failures are the few that run to the cap.}
\label{fig:analysis}
\end{figure}

\paragraph{Findings.}
\Cref{tab:ab,fig:ledger,fig:timeline,fig:analysis} give the results. The retry limit had little to cut short: under A, no session produced more than two consecutive failures, so a limit of $N = 2$ rarely met a long run, yet B stopped for help three times and doubled its interventions per success. The logs show why. Failures did not cluster: a \task{place} attempt that directly followed a failed \task{place} succeeded 8 of 13 times (62\%), against 28 of 47 (60\%) overall (\cref{fig:analysis}a); counting only retries with no human reset in between gives 6 of 10 (60\%). With so few cases this cannot rule out a moderate dependence, but we found no sign that a retry after a failure was a worse bet than any other attempt. Failures were also concentrated in \task{place}, which succeeded in 28 of 47 attempts against 30 of 32 for \task{retrieve}, and their cost was bimodal: 11 of the 19 \task{place} failures ended sooner than the median successful \task{place} (23.8\,s), 7 ran close to the 48\,s cap (\cref{fig:analysis}b).

\paragraph{What the ledger adds.}
Measured in robot minutes rather than execution seconds (\cref{fig:ledger}a), failures took 20\% of A's time and 26\% of B's, and a further 26\% and 27\% went to homing and detection, an overhead every attempt pays under either rule. Per success, the all-in cost was 50\,s of robot time for A against 57\,s for B, and 2.3\,s of human attention against 10.4\,s. B was more expensive in every currency: 15\% more robot time per success, 8\% more joint travel, and over four times as much human time. Most of that human time was not caused by the rule. Split by cause, B's 10.4\,s per success is 2.5\,s of setup at the start of a session, 7.1\,s of cubes knocked out of reach, which happens under either rule, and only 0.8\,s from two of its three retry-limit stops, the ones a person did not otherwise need to answer. A success rate would hide all of this; the ledger separates what a rule causes from what the session happened to cost.

\paragraph{An offline check of the cost-aware rule.}
Because the ledger records every attempt, the decision rule of condition D can be replayed on the logs before it runs on hardware. Using only attempts that came earlier in run order, we estimated for each task the success of an attempt straight after a same-task failure ($\hat p_{\text{now}}$) and of an attempt from a fresh start ($\hat p_{\text{fresh}}$), and asked for help only when the gain $\hat p_{\text{fresh}} - \hat p_{\text{now}}$, valued at 50\,s of robot time per success, exceeded $\lambda_h$ times the 8.8\,s a requested reset took. At all 10 decision points the estimated gain was negative (at most $-0.05$, against a break-even of $+0.17$ at $\lambda_h = 1$), so the rule never asked for help for any $\lambda_h$ from 0.05 to 4. On A's sessions it therefore makes exactly A's choices; on B's sessions it would have skipped both of B's avoidable stops. This is an offline replay, not a hardware result: it shows that on this system a cost-aware rule recovers the sensible behavior without a hand-set limit, and that it needs a setting where states differ in how likely they are to fail before it can do better than A.

\paragraph{Statistical scope.}
This is a first measurement, not a test of H1 to H3. With 3 and 2 sessions, success rates give $p = 0.21$ (Fisher's exact test, two-sided) and per-session failure-time fractions $p = 0.40$ (Mann--Whitney $U$). Including the two invalid attempts moves B to 23/34 (68\%) and changes no conclusion. Five of B's six interventions came from one session. The wrist link was re-seated after the first two sessions (one of each condition). B's success rate was lower than A's both before the repair (71\% against 81\%) and after it (60\% against 75\% and 80\%). At the session level, every A session had a higher success rate than every B session. With 3 and 2 sessions this gives a one-sided exact permutation $p = 0.1$, the smallest attainable.

\section{Limitations}
\label{sec:limitations}

The cost-aware scheduler itself has not yet been run on hardware; this paper measures the costs it is designed to trade and the baselines it must beat. The policies are frozen, so only the first regime of \cref{sec:problem} is reachable and the competence-gain objective awaits a policy that learns from its practice. Joint travel stands in for actuator wear and has not yet been related to servo current or temperature. The setting is deliberately narrow: one arm, one object, hobby-grade servos. The right weighting $\lambda$ will differ on other hardware, and the $\lambda_h$ sweep of \cref{sec:protocol} probes that only in part.

\section{Future Work}
\label{sec:future}

The ledger makes the next step direct: a scheduler that reads it. We will run the cost-aware scheduler on this arm against the baselines reported here, and then let the policies learn from their own practice, so that each attempt is chosen for what it will teach as well as what it will cost. The same ledger extends naturally to long chains of tasks and to several robots sharing one workspace, where the price of an attempt includes everything waiting behind it.

\section{Conclusion}
\label{sec:conclusion}

We built a low-cost robot that practices on its own and keeps a ledger entry for every attempt, and the ledger changed how we see autonomous practice. Three findings follow. First, practice is expensive in ways an attempt count cannot see: in the first session, more than half of the arm's execution time went to attempts that failed, and across all sessions, about a quarter of robot time went to homing and detection, an overhead every attempt pays. Second, we found no sign that failures come in streaks on this system. A retry straight after a failure succeeded as often as any other attempt, so the standard fix, asking for help after a run of failures, saved no robot time and added calls a person did not need to answer. Third, a success rate hides where the cost of practice lands. The ledger showed that most of the extra human time under the retry rule came from cubes knocked out of reach rather than from the rule, a distinction a success rate cannot make, and replaying a cost-aware rule on the same logs recovered the sensible choice without a hand-set limit. These results turn a cost that earlier methods ignored into one that can be measured, and the platform, code, data and logs are public so that others can measure it too.

\subsubsection*{Acknowledgments.}
The authors used AI assistants (including Claude, by Anthropic) for formatting, final grammar and language checks, and help with writing and reviewing the analysis and figure-generation code. The hardware, experiments, data collection and results are the authors' own; every reported number was computed from the released logs and verified by the authors, who take full responsibility for the content of this paper.


\begin{thebibliography}{99}

\bibitem{baranes2013}
Baranes, A., Oudeyer, P.-Y.: Active learning of inverse models with intrinsically motivated goal exploration in robots. Robotics and Autonomous Systems \textbf{61}(1), 49--73 (2013), \url{https://doi.org/10.1016/j.robot.2012.05.008}

\bibitem{cadene2026}
Cadene, R., Alibert, S., Capuano, F., Aractingi, M., Zouitine, A., Kooijmans, P., Choghari, J., Russi, M., Pascal, C., Palma, S., Aubakirova, D., Shukor, M., Moss, J., Soare, A., Lhoest, Q., Gallou\'edec, Q., Wolf, T.: LeRobot: An open-source library for end-to-end robot learning. In: International Conference on Learning Representations (ICLR) (2026), arXiv:2602.22818

\bibitem{eysenbach2018}
Eysenbach, B., Gu, S., Ibarz, J., Levine, S.: Leave no trace: Learning to reset for safe and autonomous reinforcement learning. In: International Conference on Learning Representations (ICLR) (2018), arXiv:1711.06782

\bibitem{fang2026}
Fang, Y., Li, Z., Tong, H., Liu, P., Chalvatzaki, G.: Find something you can't do: Agentic real-world reinforcement learning for self-improving VLA models (2026), arXiv:2609.32069

\bibitem{graves2017}
Graves, A., Bellemare, M.G., Menick, J., Munos, R., Kavukcuoglu, K.: Automated curriculum learning for neural networks. In: International Conference on Machine Learning (ICML). PMLR, vol.~70, pp. 1311--1320 (2017)

\bibitem{gupta2021}
Gupta, A., Yu, J., Zhao, T.Z., Kumar, V., Rovinsky, A., Xu, K., Devlin, T., Levine, S.: Reset-free reinforcement learning via multi-task learning: Learning dexterous manipulation behaviors without human intervention. In: IEEE International Conference on Robotics and Automation (ICRA) (2021), arXiv:2104.11203

\bibitem{gupta2023}
Gupta, A., Lynch, C., Kinman, B., Peake, G., Levine, S., Hausman, K.: Demonstration-bootstrapped autonomous practicing via multi-task reinforcement learning. In: IEEE International Conference on Robotics and Automation (ICRA) (2023), arXiv:2203.15755

\bibitem{hoque2021}
Hoque, R., Balakrishna, A., Novoseller, E., Wilcox, A., Brown, D.S., Goldberg, K.: ThriftyDAgger: Budget-aware novelty and risk gating for interactive imitation learning. In: Conference on Robot Learning (CoRL), PMLR, vol.~164, pp. 598--608 (2022), arXiv:2109.08273

\bibitem{hoque2022}
Hoque, R., Chen, L.Y., Sharma, S., Dharmarajan, K., Thananjeyan, B., Abbeel, P., Goldberg, K.: Fleet-DAgger: Interactive robot fleet learning with scalable human supervision. In: Conference on Robot Learning (CoRL), PMLR, vol.~205, pp. 368--380 (2023), arXiv:2206.14349

\bibitem{kumar2024}
Kumar, N., Silver, T., McClinton, W., Zhao, L., Proulx, S., Lozano-P\'erez, T., Kaelbling, L.P., Barry, J.: Practice makes perfect: Planning to learn skill parameter policies. In: Robotics: Science and Systems (RSS) (2024), arXiv:2402.15025

\bibitem{sharma2021}
Sharma, A., Gupta, A., Levine, S., Hausman, K., Finn, C.: Autonomous reinforcement learning via subgoal curricula. In: Advances in Neural Information Processing Systems (NeurIPS) (2021), arXiv:2107.12931

\bibitem{sharma2022}
Sharma, A., Xu, K., Sardana, N., Gupta, A., Hausman, K., Levine, S., Finn, C.: Autonomous reinforcement learning: Formalism and benchmarking. In: International Conference on Learning Representations (ICLR) (2022), arXiv:2112.09605

\bibitem{stout2010}
Stout, A., Barto, A.G.: Competence progress intrinsic motivation. In: IEEE International Conference on Development and Learning (ICDL) (2010)

\bibitem{vats2026}
Vats, S., Harithas, S., Akbulut, M.T., Raghunathan, A., Konidaris, G.: Deliberate practice: Learning robot skills under a budget (2026), arXiv:2608.13415

\bibitem{xie2022}
Xie, A., Tajwar, F., Sharma, A., Finn, C.: When to ask for help: Proactive interventions in autonomous reinforcement learning. In: Advances in Neural Information Processing Systems (NeurIPS) (2022), arXiv:2210.10765

\bibitem{yin2026}
Yin, P., Westenbroek, T., Zhang, O., Tran, J., Dagnino, I., Shilamkar, E., Mbiziwo-Tiapo, N., Bagaria, S., Liu, X., Mullins, G., Kolobov, A., Gupta, A.: Emergent dexterity via diverse resets and large-scale reinforcement learning. In: International Conference on Learning Representations (ICLR) (2026), arXiv:2603.15789

\bibitem{zhao2023}
Zhao, T.Z., Kumar, V., Levine, S., Finn, C.: Learning fine-grained bimanual manipulation with low-cost hardware. In: Robotics: Science and Systems (RSS) (2023), arXiv:2304.13705

\bibitem{zhu2020}
Zhu, H., Yu, J., Gupta, A., Shah, D., Hartikainen, K., Singh, A., Kumar, V., Levine, S.: The ingredients of real-world robotic reinforcement learning. In: International Conference on Learning Representations (ICLR) (2020), arXiv:2004.12570

\end{thebibliography}
\end{document}